\documentclass[conference]{IEEEtran}
\IEEEoverridecommandlockouts
\usepackage{cite}
\usepackage{amsmath,amssymb,amsfonts}
\usepackage{algorithmic}
\usepackage{graphicx}
\usepackage{textcomp}
\usepackage{xcolor}
\usepackage{diagbox}
\usepackage{multicol}
\usepackage{multirow}
\usepackage{array}
\usepackage{booktabs}
\def\BibTeX{{\rm B\kern-.05em{\sc i\kern-.025em b}\kern-.08em
    T\kern-.1667em\lower.7ex\hbox{E}\kern-.125emX}}
\begin{document}

\title{Dirichlet-Based Coarse-to-Fine Example Selection For Open-Set Annotation
}

\author{\IEEEauthorblockN{Ye-Wen Wang* \thanks{* Equal contribution.}, Chen-Chen Zong*, Ming-Kun Xie, Sheng-Jun Huang† \thanks{† Correspondence to: Sheng-Jun Huang.} \thanks{This work was supported by the National Key R\&D Program of China (2020AAA0107000), the Natural Science Foundation of Jiangsu Province of China (BK20222012,
BK20211517), and NSFC (62222605).}}
\IEEEauthorblockA{\textit{College of Computer Science and Technology} \\
\textit{Nanjing University of Aeronautics and Astronautics}\\
Nanjing, China \\
{linuswangg, chencz, mkxie, huangsj}@nuaa.edu.cn}

}

\maketitle

\begin{abstract}
Active learning (AL) has achieved great success by selecting the most valuable examples from unlabeled data. However, they usually deteriorate in real scenarios where open-set noise gets involved, which is studied as open-set annotation (OSA). In this paper, we owe the deterioration to the unreliable predictions arising from softmax-based translation invariance and propose a Dirichlet-based Coarse-to-Fine Example Selection (DCFS) strategy accordingly. Our method introduces simplex-based evidential deep learning (EDL) to break translation invariance and distinguish known and unknown classes by considering evidence-based data and distribution uncertainty simultaneously. Furthermore, hard known-class examples are identified by model discrepancy generated from two classifier heads, where we amplify and alleviate the model discrepancy respectively for unknown and known classes. Finally, we combine the discrepancy with uncertainties to form a two-stage strategy, selecting the most informative examples from known classes. Extensive experiments on various openness ratio datasets demonstrate that DCFS achieves state-of-art performance.
\end{abstract}


\section{Introduction}
Active learning (AL) \cite{roy2001toward, fu2013survey, huang2010active, sinha2019variational, you2014diverse, yoo2019learning, ning2021improving, hacohen2022active, fu2021agreement} is a fundamental approach aimed at alleviating considerable expenses associated with data annotation \cite{settles2009active}. It selects the most informative examples from unlabeled data and subsequently queries labels from an oracle, thereby significantly enhancing model performance at the lowest possible labeling cost \cite{huangasynchronous, mahmood2021low, ren2021survey}. Traditional AL methods typically choose examples based on either the model's prediction uncertainty or the diversity of sample features.
For example, diversity-based methods tend to choose examples based on the diversity of example features.
Uncertainty-based methods aim to select examples with the highest prediction uncertainty.
Hybrid methods consider both diversity and uncertainty, such as BADGE \cite{ash2019deep}.
\begin{figure}
	\centering
	\includegraphics[trim=40 0 0 0,width=0.45\textwidth]{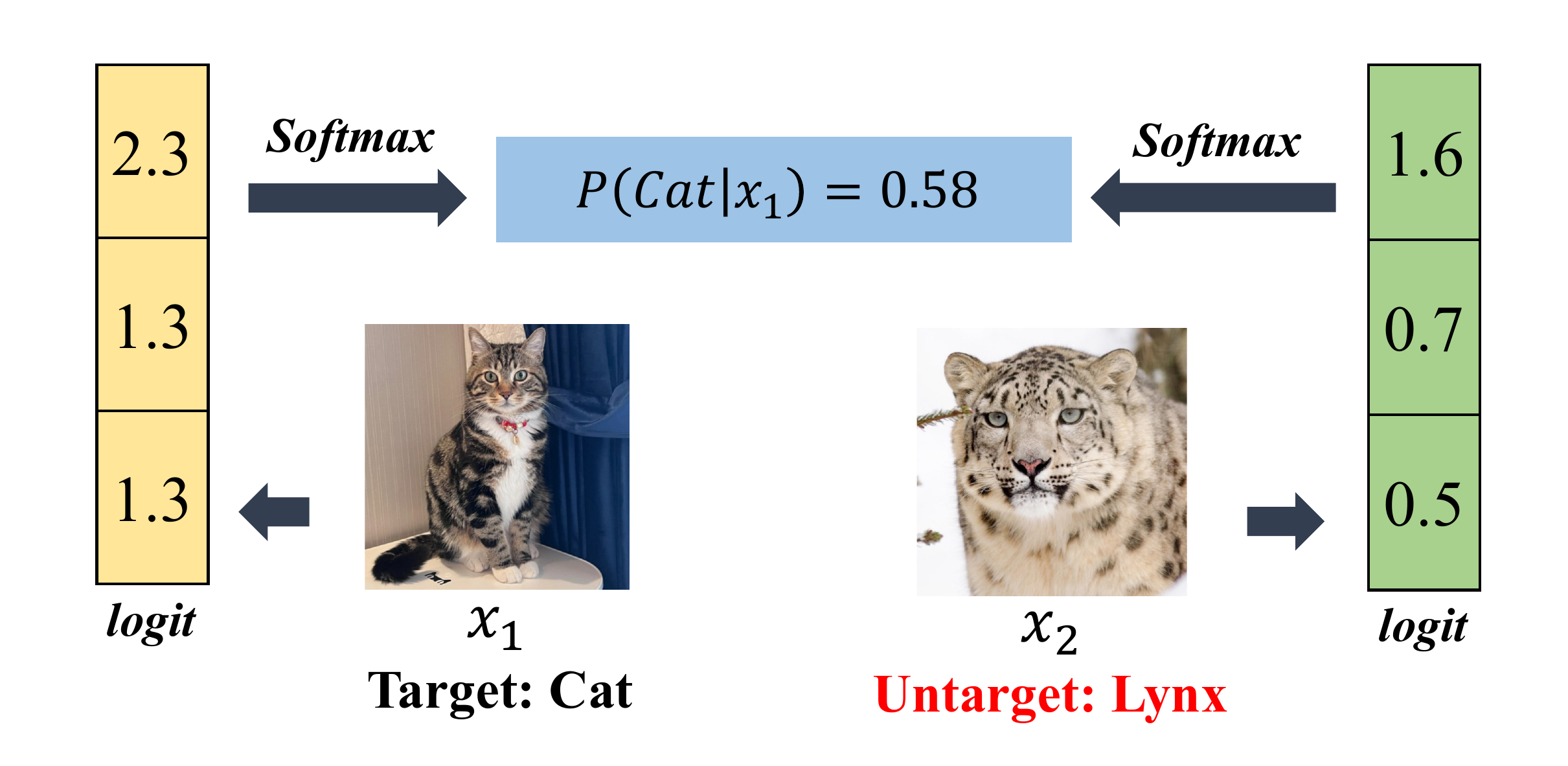}
	\caption{A possible case where softmax produces the same probabilities for known class ``Cat'' and unknown class ``Lynx''.}
	\label{fig:softmax}
  \vspace{-5mm}
\end{figure}
However, these AL methods are predominantly implemented in closed-set scenarios, where labeled and unlabeled data share the same class distribution. Unfortunately, in many real-world scenarios, since unlabeled examples are collected through cost-effective methods, \textit{e.g.}, web scraping \cite{Geng_2021}, unlabeled data inevitably contains examples not from a closed set. 
Traditional AL methods often fall short in such open-set scenarios. For example, diversity-based methods will inevitably select unknown-class examples for the diversity of their features, and the uncertainty-based methods tend to choose the unknown-class examples for the inductive biases, that the model will provide unreliable predictions for the unknown-class examples that mismatch with the distribution of training data. Manually identifying unknown classes in the pool of unlabeled data often requires a significant amount of time and financial resources. Therefore, effectively performing AL algorithms based on unlabeled data pools that contain unknown categories has become a pressing issue.

Recently, the learning scenario has been formulated as open-set annotation (OSA), and several methods have been proposed to solve this problem. These methods often query examples that have the largest similarity to labeled data \cite{du2021contrastive, park2023metaquerynet} or the max activation value (MAV) \cite{ning2022active}. 
Although these methods have made great progress in OSA, there still exist two main challenges: 

We found that all previous methods encounter the same challenge in OSA settings: the probabilities calculated by the softmax function is a point estimate that can only capture relative relationships between logits, leading to inaccuracies when evaluating unknown-class examples. 
As shown in Figure \ref{fig:softmax}, given an example $\boldsymbol{x}_1$ belongs to ``Cat'' with a logit vector of $[1.3, 2.3, 1.3]$ and another example $\boldsymbol{x}_2$ belongs to ``Lynx'' with a logit vector of $[0.7, 1.6, 0.53]$. 
Despite $\boldsymbol{x}_2$ being an unknown class ``Lynx", it will present the same probability to the category ``Cat'' by using traditional softmax-based probability. This is attributed to the translation invariance of the softmax function. Such inaccurate probabilities will misclassify unknown-class examples to known classes, severely compromising the effectiveness of example selection when conducting AL in open-set scenarios.

Moreover, despite previous OSA methods gathering more known-class examples, the lack of an informativeness measure often results in models favoring simple examples, thereby limiting their practicality on complex open-set scenarios. For example, although \cite{du2021contrastive} utilizes contrastive learning to select many known-class examples that share similar features with the labeled data, such examples promote little in model improvement for their low informativeness because the model has learned from the similar examples queried before.

\begin{figure}
	\centering
	\includegraphics[trim=40 50 0 0, width=0.45\textwidth]{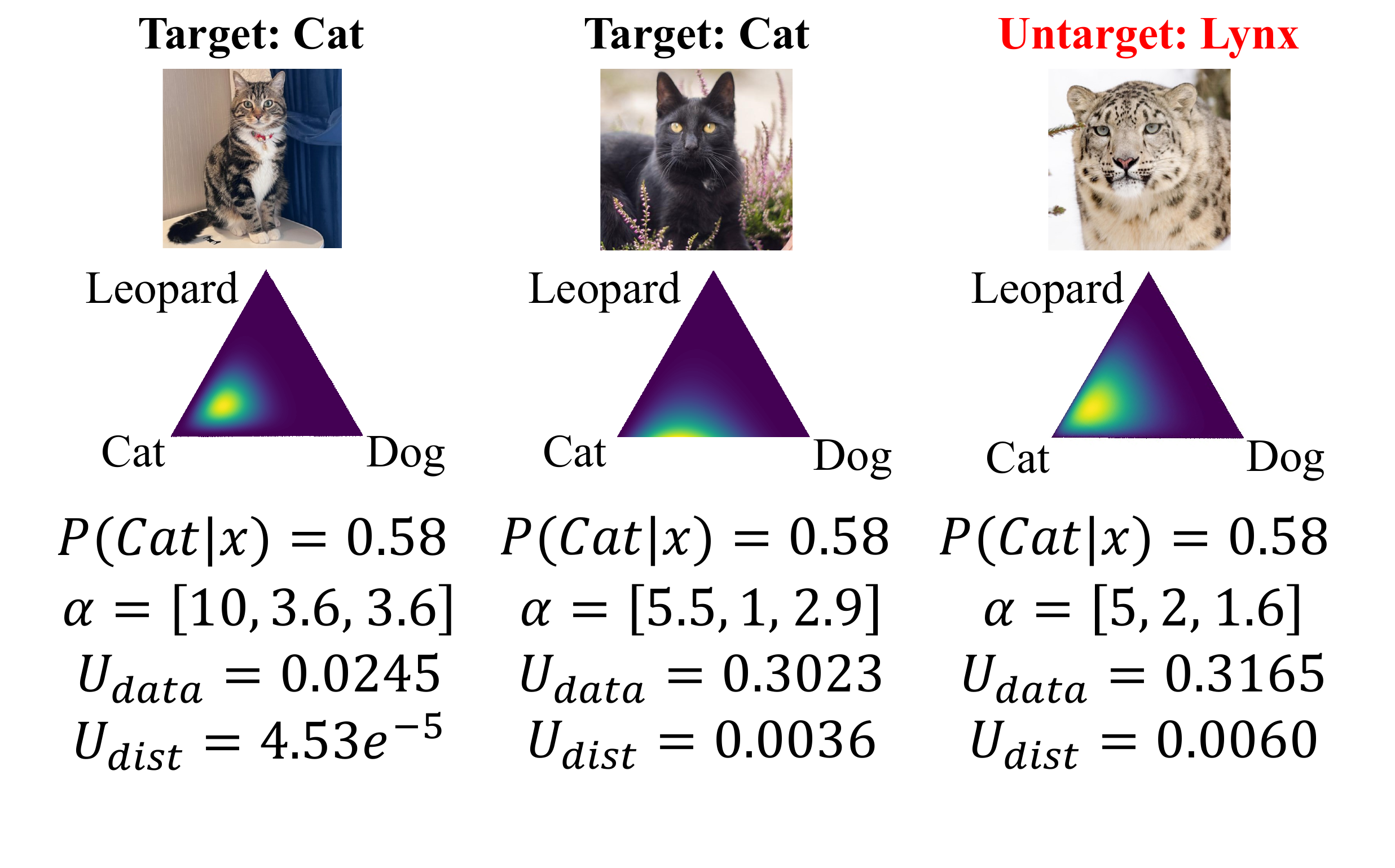}
	\caption{Dirichlet distributions, data uncertainty, and distribution uncertainty of ``Cat'' and ``Lynx'' examples with different evidence-based Dirichlet parameters that are exponentially correlated to logits.} 
	\label{fig:edl}
 \vspace{-5mm}
\end{figure}

We attribute the first challenge to the translation invariance caused by the softmax function. Inspired by evidential deep learning (EDL), which treats the model's predictions as a random variable rather than a simple point estimate, we further refine the logit outputs of the model by imposing a Dirichlet prior on the predicted probabilities. For example, as illustrated in Figure \ref{fig:edl}, although ``Cat" and ``Lynx" share the same prediction on ``Cat", the prediction with a higher evidence value of $10$ generates a more concentrated distribution closer to the corner, while the unknown ``Lynx" with lower evidence value of $5$ generates a more dispersed distribution. This implies that, compared to softmax-based predictions, EDL can break its translation invariance and present more reliable predictions.

Furthermore, to balance the informativeness and purity of examples in querying, we use the mean and variance of the distribution entropy of the simplex as measures of data and distribution uncertainty. Normal known-class examples, like ``Cat", located close to the corner of simplex, exhibit lower uncertainty, while unknown-class examples, like ``Lynx", located farther from the corner, show higher uncertainty. Rare examples between two classes, like the black ``Cat", demonstrate higher distribution uncertainty. This suggests that we can differentiate informative known-class examples based on these two metrics.

Based on this, we propose a Dirichlet-based Coarse-to-Fine Example Selection (DCFS) strategy to select informative known-class examples. 
To break softmax's translation invariance, we train the target model using simplex-based evidential deep learning (EDL) which treats the model's predictions as a random variable with the Dirichlet prior. To query examples that are both informative and known-class ones, we decouple the probabilistic simplex into two different uncertainty metrics, i.e., distribution uncertainty for purity measurement and data uncertainty for informative measurement. Then, to further distinguish between hard known-class examples and unknown-class ones, inspired by \cite{peng2021alforlane}, we introduce a model discrepancy score by implementing two classifier heads atop the target model and fine-tuning the model on unlabeled data. An example with a large discrepancy score will more likely belong to an unknown class.
Extensive experiments demonstrate that DCFS not only outperforms previous methods on the prediction accuracy across all openness ratios, but also selects more informative examples instead of simple known-class examples, substantially improving model performance and achieving state-of-the-art performance. 

\begin{figure*}
	\centering
	\includegraphics[trim=100 60 60 0, width=0.8\textwidth]{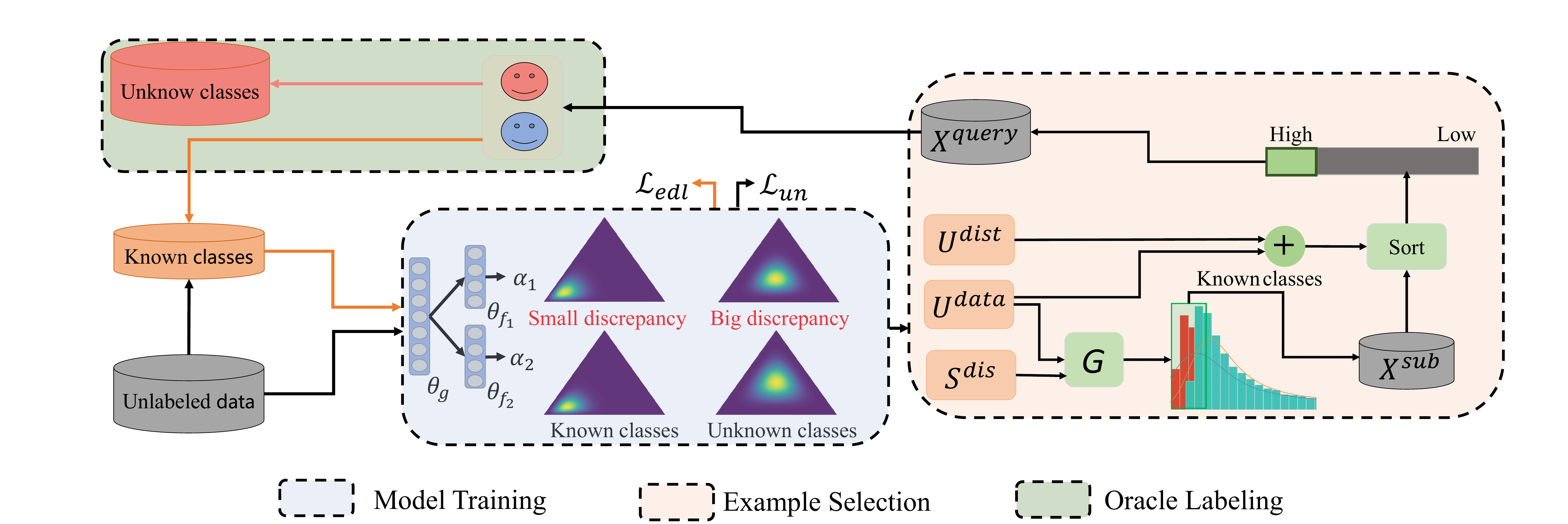}
	\caption{The overall procedure of our proposed approach. In the model training phase, the discrepancy score $S^{dis}$ of the two classifier heads is amplified first with $\mathcal{L}_{un}$ and then we train the model with an evidential-based loss $\mathcal{L}_{edl}$. During the example selection stage, we select informative known-class examples with a combination of data uncertainty $U^{data}$, distribution uncertainty $U^{dist}$, and discrepancy score $S^{dis}$. Lastly, the selected $X^{query}$ is sent for oracle labeling.}
	\label{fig:model}
 \vspace{-5mm}
\end{figure*}

\section{Methodology}
\label{sec:methodology}
\subsection{Preliminary}
\noindent\textbf{Notations.} In open-set annotation (OSA) problems, a limited labeled data pool $\mathcal{D}_{L} = \{(\boldsymbol{x}_i, y_i)\}_{i=1}^{N_L}$ each $\boldsymbol{x}_i \in \mathcal{X}$ belongs to one of $C$ known classes $\mathcal{Y} = \{c\}_{c=1}^C$ is provided, along with an unlabeled data pool $\mathcal{D}_{U} = \{x_i\}_{i=1}^{N_U}$ which is a mixture of both known and unknown class examples. $r$ denotes the openness ratio and represents the proportion of unknown-class examples occupying. In each iteration, active learning (AL) algorithms query $b$ examples from $\mathcal{D}_{U}$ to form the query set $X^{query}$, which is then sent to the oracle for annotations and updated to the corresponding data pool.

\noindent\textbf{Overview.} The framework of our proposed approach is illustrated in Figure \ref{fig:model}, which primarily involves three stages: model training, example selection, and oracle labeling. 
In the Model Training phase, two classifier heads are implemented atop the initial model. We first train the target model with simplex-based evidential deep learning (EDL) to collect evidence and produce simplex-based distribution for examples. Then, we train the model to amplify the model discrepancy of the two classifier heads for unlabeled data. These two training stages are processed alternately, leading that known-class examples own lower model discrepancy and a more close-to-corner simplex-based distribution. 
In the Example Selection phase, we extract the data uncertainty and distribution uncertainty from the distributed entropy of the simplex. These two uncertainties are combined with the model discrepancy to estimate whether examples are from known classes and informative accordingly.
In the Oracle Labeling phase, we send the queried examples to annotators for class labels. Queried known-class examples will be updated to the labeled data. 

\subsection{Simplex-Based Evidential Training Methodology}
\label{EDL}
As discussed in the Introduction, one distinct challenge in OSA is misclassifying unknown-class examples as known-class ones due to the softmax's translation invariance. To counter this, we propose to introduce Simplex-based evidential deep learning (EDL), which is based on Subjective Logic \cite{josang2016subjective} and Dempster–Shafer Theory \cite{shafer1992dempster} to collect the evidence of examples being every category. 
Compared to the softmax-based prediction, EDL serves the exponential of logits as evidence and represents the evidence for all categories as the parameter vector $\boldsymbol{\alpha}$ of Dirichlet prior. Through this, a distribution of probability $\boldsymbol{\rho}$ for a given example can be described on a class simplex $\Delta^C = \{\boldsymbol{\rho} | \sum_{c=1}^C\rho_c=1\}$. 

Formally, EDL treats the probability $\boldsymbol{\rho}$ as a random variable and calculates it for each input $\boldsymbol{x}_i$ based on the Dirichlet prior. With $\boldsymbol{\theta}$ denoting the model parameters, we compute the evidence vector $\boldsymbol{\alpha}_i$ for a given example $\boldsymbol{x}_i$ as $\boldsymbol{\alpha}_i = g(f(\boldsymbol{x}_i, \boldsymbol{\theta}))$, where $g(\cdot)$ represents an exponential function to guarantee that evidence collected by EDL is positive. Then, the probability density of $\boldsymbol{\rho}_i$ can be calculated by:
\begin{equation}
    \label{eq1}
    p(\boldsymbol{\rho}_i|\boldsymbol{x}_i, \boldsymbol{\theta}) = Dir(\boldsymbol{\rho}_i|\boldsymbol{\alpha}_i) = 
        \frac{\Gamma(\sum_{c=1}^C\alpha_{ic})}{\prod_{c=1}^C\Gamma(\alpha_{ic})}\prod_{c=1}^C\rho_{c}^{\alpha_{ic}-1}.
\end{equation}
By marginalizing over $\boldsymbol{\rho}_i$, the predicted probability on a given class $c$ can be obtained by:
\begin{equation}
\label{eq2}
\begin{split}
    P(y=c|\boldsymbol{x}_i, \boldsymbol{\theta}) &= \int p(y=c|\boldsymbol{\rho}_i)p(\boldsymbol{\rho}_i, \boldsymbol{\theta})d\boldsymbol{\rho}_i \\
    &= \frac{g(f_c(\boldsymbol{x}_i,\boldsymbol{\theta}))}{\sum_{k=1}^C g(f_k(\boldsymbol{x}_i,\boldsymbol{\theta})}.
\end{split}
\end{equation}
To produce a sharp Dirichlet distribution situated at the corner for known class examples, we introduce the loss $\mathcal{L}_{nll}$ to minimize the negative logarithm of the marginal likelihood for all labeled data by:
\begin{equation}
\label{eq3}
\begin{split}
    \mathcal{L}_{nll} &= -\frac{1}{N}\sum_{i=1}^{N}log[P(y=c|\boldsymbol{\rho}_i)]p(\boldsymbol{\rho}_i|\boldsymbol{x}_i, \boldsymbol{\theta}) \\
    &= \frac{1}{N}\sum_{i=1}^{N}\sum_{c=1}^C Y_{ic}[log(\sum_{j=1}^C \alpha_{ij})-log \alpha_{ic}],
\end{split}
\end{equation}
where $Y_{ic}$ is the $c$-th element of the one-hot label vector $\boldsymbol{Y}_i$ for $\boldsymbol{x}_i$ with the scalar label $y_i$. 
To reduce the evidence collected on complementary labels (labels other than the given label), 
we reduce the evidence on other categories to zero by implementing the Kullback-Leibler Divergence loss $\mathcal{L}_{kl}$ defined as:
\begin{equation}
\label{eq4}
    \begin{split}
        \mathcal{L}_{kl} &= \frac{1}{NC}\sum_{i=1}^{N}\sum_{c=1}^{C}D_{KL}(Dir(\boldsymbol{\rho}|\Tilde{\boldsymbol{\alpha}}_i)||Dir(\boldsymbol{\rho}|\mathbf{1})) \\ 
        &= \frac{1}{N}\sum_{i=1}^{N}[log[\frac{\Gamma(\sum_{j=1}^C)\alpha_{ij}}{\Gamma(C)\prod_{j=1}^{C}\Gamma(\alpha_{ij})}] \\
        &+ \sum_{c=1}^{C}(\Tilde{\alpha}_{ic}-1)[\Phi(\Tilde{\alpha}_{ic})-\Phi(\sum_{j=1}^{C}\Tilde{\alpha}_{ij})]],
    \end{split}
\end{equation}
where $D_{KL}$ denotes the Kullback-Leibler Divergence (KLD), $\Gamma$ represents the Gamma function, $\Phi$ indicates the digamma function, $\odot$ is the element-wise multiplication, and $\Tilde{\boldsymbol{\alpha}}_i = \boldsymbol{Y}_i + (1 - \boldsymbol{Y}_i) \odot \boldsymbol{\alpha}_i$ is a temp form by removing the given label evidence from the original $\boldsymbol{\alpha}_i$. Finally, the overall EDL loss $\mathcal{L}_{edl}$ is formulated as:
\begin{equation}
\label{eq5}
    \mathcal{L}_{edl} = \mathcal{L}_{nll} + \mathcal{L}_{kl}.
\end{equation}
Training with this will lead the ``Cat'' in Figure \ref{fig:edl} to collect more evidence of being ``Cat'' and less evidence of other categories, generating a more concentrated distribution near the corner than the unknown ``Lynx'', which is superior to softmax-based metric on distinguishing between them by the simplex-based distribution and thus can resolve the translation invariance challenge.

\subsection{Evidence-Based Data and Distribution Uncertainty}
\label{Unc}
The known-class examples usually own a sharp distribution near the corner of the simplex, while the unknown-class ones own those in the center. Inspired by \cite{xie2023dirichletbased}, we model such correlations between examples and distributions by splitting expected prediction entropy into two parts: data and distribution uncertainty. Here, data uncertainty indicates the probability of examples being known or unknown class. Distribution uncertainty represents the degree of evidence collected during the training stage.

Specifically, data uncertainty is the mean of the expected entropy for all possible $\boldsymbol{\rho}$ and is defined as:
\begin{equation}
\label{eq6}
    \begin{split}
        {U}^{data}(\boldsymbol{x}_i, \boldsymbol{\theta}) &\triangleq \mathbb{E}_{p(\boldsymbol{\rho}|\boldsymbol{x}_i,\boldsymbol{\theta})}[H[P(y|\boldsymbol{\rho})]] \\
        &= \sum_{c=1}^C \bar{\rho}_{jc}(\varphi(\sum_{k=1}^C \alpha_{ik}+1)-\varphi(\alpha_{ic}+1)).
    \end{split}
\end{equation}
where the expected entropy is $H[P(y|\boldsymbol{\rho})]$. An example with higher data uncertainty focuses more concentrations on the center of simplex where entropy is larger, and can be characterized as unknown-class accordingly. The distribution uncertainty is the variance of the expected entropy and can be used to evaluate the degree of evidence collection. Formally, we evaluate it by computing evidence to be each category using the mutual information $I[y,\boldsymbol{\rho}|\boldsymbol{x}_i,\boldsymbol{\theta}]$ as follows:
\begin{equation}
\label{eq7}
    \begin{split}
        &= \sum_{c=1}^C\bar{\rho}_{ic}(\varphi(\alpha_{ic}+1)-\varphi(\sum_{k=1}^C\alpha_{ik}+1)) -\sum_{c=1}^C\bar{\rho}_{ic}\log \bar{\rho}_{ic}.
    \end{split}
\end{equation}
An example with higher distribution uncertainty indicates that the model is unsure whether the example was learned before, which can be seen as an informativeness data point.

\begin{table*}[ht]
	\small
	\centering
\caption{ Final test accuracy comparasion on CIFAR10, CIFAR100, and Tiny-ImageNet. The best performances are highlighted in bold.}
	{\begin{minipage}{0.9\linewidth}
			\centering
			\resizebox{1\textwidth}{!}{
				\setlength{\tabcolsep}{1.5mm}{
					\begin{tabular}{ccccccccccccc}
						\toprule
						
					Dataset	& & \multicolumn{3}{c}{CIFAR10} & & \multicolumn{3}{c}{CIFAR100} & & \multicolumn{3}{c}{Tiny-ImageNet} \\
						\cmidrule{1-1} \cmidrule{3-5}\cmidrule{7-9}\cmidrule{11-13}
						\diagbox[]{Method}{Ratio} & & 0.2 & 0.4 & 0.6 & & 0.2 & 0.4 & 0.6 & & 0.2 & 0.4 & 0.6\\
						\cmidrule{1-1} \cmidrule{3-5}\cmidrule{7-9}\cmidrule{11-13}
						Random & & 80.5$\pm$0.6 & 79.8$\pm$0.4 & 85.2$\pm$0.9 & & 52.7$\pm$0.6 & 53.9$\pm$0.8 & 55.1$\pm$0.2 & & 42.5$\pm$0.2 & 43.8$\pm$0.7 & 47.6$\pm$0.3\\
						Coreset & & 80.8$\pm$1.2 & 78.9$\pm$0.6 & 84.7$\pm$0.6 & & 55.2$\pm$1.0 & 56.2$\pm$0.6 & 57.5$\pm$1.5 & & 42.1$\pm$0.5 &  43.2$\pm$0.6 &  47.2$\pm$0.3\\
						LC & & 82.0$\pm$0.3 & 78.3$\pm$0.5 & 85.2$\pm$0.5 & & 51.6$\pm$0.5 & 51.4$\pm$0.9 & 50.1$\pm$1.2 & & 41.1$\pm$0.3 &  42.3$\pm$0.4 &  44.8$\pm$0.6\\
						Entropy & & 81.8$\pm$0.3 & 75.3$\pm$0.7 & 86.1$\pm$0.7 & & 50.9$\pm$1.5 & 50.4$\pm$1.4 & 50.1$\pm$2.2 & & 40.6$\pm$0.2 & 41.8$\pm$0.8 &  44.4$\pm$0.6\\
						Margin & & 82.5$\pm$0.5 & 80.6$\pm$0.3 & 86.8$\pm$0.3 & & 54.2$\pm$0.5 & 55.0$\pm$0.5 & 53.6$\pm$0.5& & 41.4$\pm$2.2 &  43.6$\pm$0.2 &  47.1$\pm$1.1\\
                            BADGE & & 83.6$\pm$0.6 & 81.5$\pm$0.3 & 84.9$\pm$2.1 & & 55.8$\pm$0.5 & 55.9$\pm$0.5 & 55.5$\pm$0.6& & 42.1$\pm$0.5 &  43.2$\pm$0.6 &  47.2$\pm$0.7\\
						CCAL & & 77.7$\pm$0.1 & 78.2$\pm$0.2 & 85.3$\pm$0.6 & & 49.6$\pm$0.1 & 56.1$\pm$0.2 & 60.4$\pm$0.6 & & 41.0$\pm$0.3 &  42.7$\pm$0.3 &  46.9$\pm$0.5\\
						LfOSA & & 70.5$\pm$0.6 & 74.1$\pm$0.4 & 82.8$\pm$0.5 & & 48.2$\pm$1.0 & 52.1$\pm$0.4 & 57.3$\pm$0.2 & & 39.3$\pm$0.7 & 42.7$\pm$0.3 &  48.1$\pm$1.0\\
                            \midrule
						\textbf{DCFS(ours)} & & \textbf{84.8$\pm$0.8} & \textbf{83.2$\pm$0.8} & \textbf{86.8$\pm$0.2}& & \textbf{57.7$\pm$0.3} & \textbf{59.9$\pm$0.8} & \textbf{60.4$\pm$0.5} & & \textbf{42.8$\pm$0.4} &  \textbf{45.3$\pm$0.4} &  \textbf{50.5$\pm$0.7}\\
						\bottomrule
				\end{tabular}}
			}
	\end{minipage}}
        
        \label{table:main_exp}
        \vspace{-5mm}
\end{table*}

\subsection{Discrepancy-Based Training Methodology}
\label{differentiation}
Due to data scarcity limiting the collection of evidence, depending solely on $U^{data}$ may still result in the misclassification of hard known-class examples as unknown-class ones. Previous work shows \cite{peng2021alforlane} that ``Noisy" examples tend to exhibit high model discrepancy when models are trained in different manners. We extend this concept by considering unknown-class examples as noisy ones and retaining the hard-known class with lower model discrepancy. In practice, we implement two classifier heads atop the model to produce prediction discrepancy, which forms the discrepancy score aiding in unknown-class filtering. 

Formally, let $\boldsymbol{\theta}_g$ represents the parameters of feature representation, $\boldsymbol{\theta}_{f_1}$ and $\boldsymbol{\theta}_{f_2}$ denotes the parameters from two classifier heads respectively.
$\boldsymbol{\alpha}_1$ and $\boldsymbol{\alpha}_2$ are the evidence outputs for the given example $\boldsymbol{x}_i$. 
Specifically, after the model has been trained on the labeled data pool, we freeze the parameters $\boldsymbol{\theta}_{f_1}$ and $\boldsymbol{\theta}_{f_2}$, and only fine-tune the parameters $\boldsymbol{\theta}_g$ on unlabeled data pool. In practice, we minimize the two-head prediction discrepancy with the Jensen-Shannon Divergence (JSD) $D_{JSD}$ by loss $\mathcal{L}_{close}$ defined as:
\begin{equation}
\label{eq8}
    \begin{split}
        {\arg\min}_{\boldsymbol{\theta}_g}&\mathcal{L}_{close}(\boldsymbol{\alpha}_1, \boldsymbol{\alpha}_2)=\sum_{i=1}^{N}w_i D_{JSD}(\boldsymbol{\alpha}^{(1)}_i || \boldsymbol{\alpha}^{(2)}_i) \\
        &w_i = \frac{1}{N}(1-\sigma(U^{data}_i-\tau_1)).
    \end{split}
\end{equation}
Here, data with low $U^{data}$ will be assigned a larger update weight. Then, we fix the parameters $\boldsymbol{\theta}_g$, and only fine-tune the parameters $\boldsymbol{\theta}_{f_1}$ and $\boldsymbol{\theta}_{f_2}$ same on unlabeled data pool. This time, we amplify the two-head prediction discrepancy for high $U^{dist}$ examples by: 
\begin{equation}
\label{eq9}
    \begin{split}
        {\arg\min}_{\boldsymbol{\theta}_{f1}, \boldsymbol{\theta}_{f2}}&\mathcal{L}_{dis}(\boldsymbol{\alpha}_1, \boldsymbol{\alpha}_2)=\sum_{i=1}^{N}w_i(1- D_{JSD}(\boldsymbol{\alpha}^{(1)}_i || \boldsymbol{\alpha}^{(2)}_i)) \\
        &w_i = \frac{1}{N}\sigma(U^{dist}_i-\tau_2).
    \end{split}
\end{equation}
Through this, data in $\mathcal{D}_U$ will exhibit two distinct simplex-based distributions based on the outputs from the two classifier heads. Here, we define the L2-norm minus between outputs $\boldsymbol{\alpha}_1$ and $\boldsymbol{\alpha}_2$ as the discrepancy score $S^{dis}$ and have it participate into the known-class example selection:
\begin{equation}
\label{eqcls_un}
    S^{dis}(\boldsymbol{x}_i;\boldsymbol{\alpha}^{(1)}_i, \boldsymbol{\alpha}^{(2)}_i) = \sqrt{\sum_{c=1}^{C}({\alpha}^{(1)}_{ic}-{\alpha}^{(2)}_{ic}))^2}
\end{equation}

\subsection{Dirichlet-Based Coarse-to-Fine Selection Strategy}
\label{sampling strategy}
Finally, by integrating three uncertainty measurements, we propose a coarse-to-fine strategy to gradually identify informative known-class examples.
In the coarse stage, considering that known-class examples usually present low data uncertainty and model discrepancy, we combine these two scores as $S^{dis}+\alpha U^{data}$. Then, we fit a two-mode Gaussian Mixture Model (GMM) based on the score to filter out unknown class examples.
By denoting $G(S^{dis}+\alpha U^{data})$ as the probability belonging to known-class mode, we adopt a threshold $\lambda$ to select known class examples as
\begin{equation} 
\label{eq11}
    X^{sub} = \cup_{\boldsymbol{x} \in \mathcal{D}_{U}}[G(S^{dis}(\boldsymbol{x}) + \alpha U^{data}(\boldsymbol{x}))> \lambda]
\end{equation}
where $X^{sub}$ is the subset of unlabeled data selected in the coarse stage.
Moreover, as distribution uncertainty indicates the informativeness of examples and data uncertainty indicates the dissimilarity of the given example to labeled ones, we select the informative and representative examples based on the measurement $\beta U^{data} + U^{dist}$. The selection procedure can be formulated as:
\begin{equation}  
\label{eq12}
    X^{query} = {\arg\max}_{\boldsymbol{x} \in X^{sub}}(\beta U^{data}(\boldsymbol{x}) + U^{dist}(\boldsymbol{x}))
\end{equation}

\section{Experiments}
\label{sec:Experiments}

\subsection{Experimental Setting}
\noindent\textbf{Datasets.} The experiments are conducted on three typical benchmark datasets: CIFAR10, CIFAR100 \cite{krizhevsky2009learning}, and Tiny-ImageNet \cite{yao2015tiny}. CIFAR10 and CIFAR100 consist of 50,000 training images and 10,000 test images. Tiny-ImageNet is a subset of the ImageNet, which contains 200 classes with 500 training images per class. For all datasets, we experiment with the openness ratio $r$ at 0.2, 0.4, and 0.6. We randomly select $1\%$, $8\%$, and $8\%$ of known-class examples as the initial labeled data for CIFAR10, CIFAR100, and Tiny-ImageNet.

\noindent\textbf{Implementation Details.} We use ResNet-18 \cite{he2016deep} as the backbone network for all datasets and add two classifier heads atop. We conduct six cycles of AL queries for each experiment and train the model for 100 epochs using SGD as the optimizer, where momentum is 0.9, weight decay is 1e-4, and batch size is 128. The learning rate is 0.01 initially and reduces by a factor of 10 at epochs 60 and 80. The query size $b$ is 1500 for CIFAR10 and CIFAR100, and 3000 for Tiny-ImageNet. The hyperparameters $\tau_1$ is 7.0, $\tau_2$ is $-5.0$, threshold $\lambda$ is 0.5, $\alpha$ is 1.0 and $\beta$ is 0.5 for all datasets. We repeat all the experiments three times and report the average performance. 

\noindent\textbf{Comparing Methods.} In our experiments, we compare with the following methods:
\begin{enumerate}
    \item Random: it selects examples randomly from unlabeled data.
    \item Uncertainty metrics include least confidence (LC) \cite{li2006confidence}, entropy \cite{holub2008entropy} and margin \cite{balcan2007margin}: such methods select examples with highest softmax-based uncertain.
    \item Diversity metric Coreset \cite{sener2017active}: it selects examples considering the diversity of example features.
    \item Hybrid metric BADGE \cite{ash2019deep}: it combines both uncertainty and diversity to select examples.
    \item State-of-the-art OSA metrics CCAL \cite{du2021contrastive} and LfOSA \cite{ning2022active}: they use additional costs such as contrastive learning and another model to detect known-class examples.
\end{enumerate}

\subsection{Performance Comparison}
Table \ref{table:main_exp} shows the final round test accuracy of our proposed DCFS and compared methods on  CIFAR10, CIFAR100, and Tiny-ImageNet. DCFS consistently outperforms all compared methods across all openness ratios. The performance highlights the robustness and effectiveness of our method. Furthermore, we can also observe that: 
1) Traditional AL methods demonstrate effectiveness in scenarios with a high openness ratio, akin to close-set scenarios. However, their performance degrades as the openness ratio increases. This observed phenomenon aligns with our motivation, as increased openness ratios are associated with heightened uncertainty and distribution disparities, inevitably introducing unknown-class examples.
2) The performances of state-of-the-art OSA methods CCAL and LfOSA deteriorate as the openness ratio decreases. The experimental results indicate that CCAL and LfOSA are ineffective in handling complex open-set scenarios, especially when the openness ratio is agnostic in real life. This ineffectiveness arises from their inclination to prioritize the selection of known-class examples over informative ones, causing difficulties in exploring known-class examples dissimilar to the labeled data. 
3) In contrast, our DCFS method achieves an optimal balance between known-class examples and informative examples across all openness ratios.

\begin{figure}
	\centering
	\includegraphics[trim=0 20 0 0, width=0.4\textwidth]{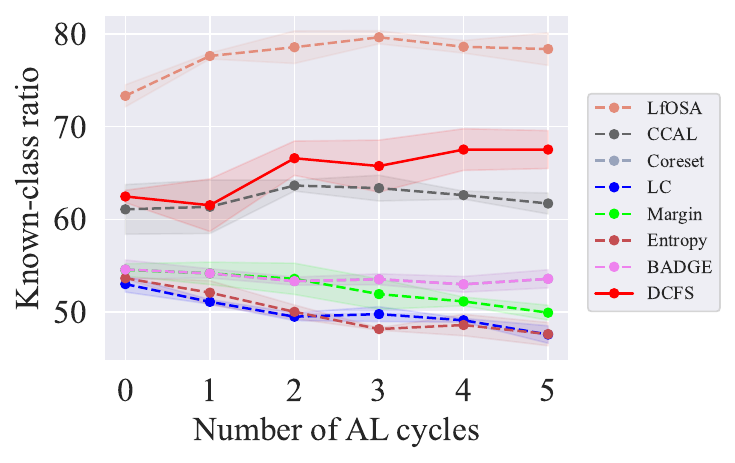}
	\caption{The curve of select accuracy of each cycle under CIFAR10 with 0.4 openness ratio.} 
	\label{fig:select_acc}
\end{figure}

\begin{figure}
	\centering
	\includegraphics[trim=0 120 0 80, width=0.5\textwidth]{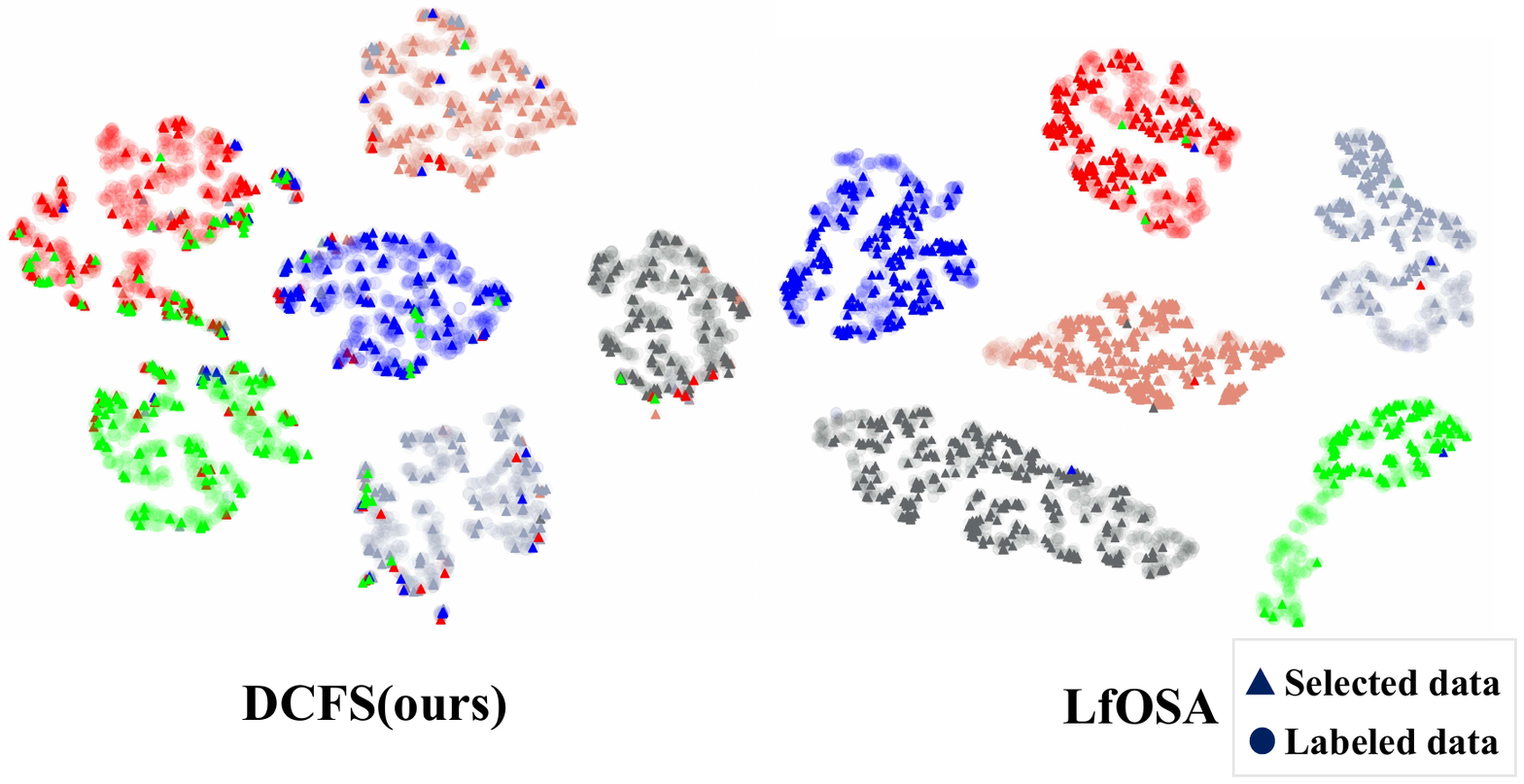}
	\caption{The t-NSE graph of selected known-class examples and existing labeled examples under CIFAR10 with 0.4 openness ratio.} 
	\label{fig:nse}
  \vspace{-5mm}
\end{figure}

To explain why LfOSA and CCAL fail in a high openness ratio detailedly, we plot the curve of query precision of each cycle and the t-NSE distribution of selected examples and labeled data. Figure \ref{fig:select_acc} shows that DCFS outperforms all compared methods in the detection of known-class examples, except for LfOSA. Nevertheless, as illustrated by the t-NSE graph in Figure \ref{fig:nse}, LfOSA tends to select examples similar to labeled data, prioritizing the detection of known-class examples while neglecting the informativeness of examples. 

\begin{table}[t]
		\centering
          \caption{ The average performance of ablation study, \textbf{w/o} is denoted as the removal of Discrepancy Score $S^{dis}$ and EDL. }
        \begin{tabular}{lcccc}
           \toprule
            
            Dataset	& & CIFAR10 & & CIFAR100 \\
            \cmidrule{1-1} \cmidrule{3-3}\cmidrule{5-5}
                DCFS $\bf{w}\backslash o$ $S^{dis}$ & & 84.5 & & 58.2\\ 
                DCFS $\bf{w}\backslash o$ EDL & & 83.7 & & 54.6\\
                \midrule
            \textbf{DCFS} & & \textbf{85.0} & & \textbf{59.3}\\
            \bottomrule
        \end{tabular}

        \label{table:snetv2}
 \vspace{-5mm}
\end{table}

\subsection{Ablation Study}
We ablate on two main components of DCFS on CIFAR10 and CIFAR100 to demonstrate their effectiveness: 

\noindent\textbf{Effect of $S^{dis}$.} To study the effect $S^{dis}$ originated from the two classifier heads, we remove one of them and use only $U^{data}$ in the coarse stage. From Table \ref{table:snetv2}, the removal of classifier head and $S^{dis}$ leads to performance degrading, suggesting that hard known-class examples do exist and $S^{dis}$ can retain them.

\noindent\textbf{Effect of EDL.} As shown in Table \ref{table:snetv2}, the performance of DCFS which is trained by Cross-Entropy loss\cite{RUBINSTEIN199789} decreases significantly on all datasets, indicating the damage of softmax's translation invariance and robustness of EDL in selecting examples. However, our DCFS with Cross-Entropy loss still outperforms all other methods, demonstrating the informativeness of our proposed query strategy.

\section{Conclusion}
\label{sec:majhead}
In this paper, we propose a \textbf{D}irichelet-based \textbf{C}oarse-to-\textbf{F}ine Example \textbf{S}election (DCFS) strategy that selects known-class informative examples in open-set data. On the one hand, DCFS collects evidence for each class using evidential deep learning (EDL), with split evidence-based data and distribution uncertainty depicting the correctness and evidence level of the examples.
On the other hand, two classifier heads are implemented atop the model to retain hard known-class examples by evaluating the discrepancy score originating from amplified model discrepancy. Finally, informative known-class examples are selected with a strategy combined with the model discrepancy, data, and distribution uncertainty. Experimental results demonstrate that our approach achieves state-of-the-art performance in intricate open-set scenarios across all settings.

\bibliographystyle{IEEEtran}
\bibliography{icme}

\end{document}